\pdfoutput=1

\documentclass[11pt]{article}

\usepackage{acl}
\usepackage{times}
\usepackage{latexsym}
\usepackage[T1]{fontenc}
\usepackage[utf8]{inputenc}
\usepackage{microtype}
\usepackage{inconsolata}

\usepackage{xspace}
\usepackage{graphicx}
\usepackage{amsmath}
\usepackage{tabularx}
\usepackage{enumitem}
\usepackage{mdframed}
\usepackage{xcolor,colortbl}
\usepackage{url}
\usepackage{booktabs}
\usepackage{booktabs}       
\usepackage{amsfonts}       
\usepackage{hyperref}

\newmdenv[
  backgroundcolor=red!05,
  linecolor=quoteborder,
  skipabove=1em,
  skipbelow=0em,
  leftline=true,
  topline=false,
  bottomline=false,
  rightline=false,
  linecolor=red!66,
  linewidth=4pt
]{githubquote}

\newcommand{\ourmethod}{\textsc{R2S}}
\newcommand{\gmodel}{\textsc{gLLM}}
\newcommand{\instruct}{\textsc{gInstruct}}
\newcommand{\benchmark}{\textsc{k-Bench}}
\newcommand{\cod}{CoD}

\title{Raw Text is All you Need: Knowledge-intensive Multi-turn Instruction Tuning for Large Language Model}

\author{
  Xia HOU\textsuperscript{\rm 1},
  Qifeng Li\textsuperscript{\rm 1},
  Jian Yang\textsuperscript{\rm 2}\thanks{\ \ Corresponding Author.},
  Tongliang Li\textsuperscript{\rm 1}, 
  Linzheng Chai\textsuperscript{\rm 2}, \\
  {\bf Xianjie Wu}\textsuperscript{\rm 2}, 
  {\bf Hangyuan Ji}\textsuperscript{\rm 2},
  {\bf Zhoujun Li}\textsuperscript{\rm 2}, 
  {\bf Jixuan Nie}\textsuperscript{\rm 1},
  {\bf Jingbo Dun}\textsuperscript{\rm 1},
  {\bf Wenfeng Song}\textsuperscript{\rm 1},\\
  \textsuperscript{\rm 1}Computer School, Beijing Information Science \& Technology University; \\
  \textsuperscript{\rm 2}State Key Laboratory of 
Complex \& Critical Software Environment, Beihang University \\
  \{houxia, liqifeng, tonyliangli, niejixuan, dunjingbo, songwenfeng\}@bistu.edu.cn; \\
  \{jiaya, challenging, wuxianjie, jhy\_1, lizj\}@buaa.edu.cn \\
}

\begin{document}
\maketitle

\renewcommand{\thefootnote}{\fnsymbol{footnote}} 
\renewcommand{\thefootnote}{\arabic{footnote}} 

\begin{abstract}
Instruction tuning as an effective technique aligns the outputs of large language models (LLMs) with human preference. But how to generate the seasonal multi-turn dialogues from raw documents for instruction tuning still requires further exploration.
In this paper, we present a novel framework named \ourmethod{} that leverages the \cod{}—Chain of Dialogue logic to guide large language models (LLMs) in generating knowledge-intensive multi-turn dialogues for instruction tuning. By integrating raw documents from both open-source datasets and domain-specific web-crawled documents into a benchmark \benchmark{}, we cover diverse areas such as Wikipedia (English), Science (Chinese), and Artifacts (Chinese). Our approach first decides the logic flow of the current dialogue and then prompts LLMs to produce key phrases for sourcing relevant response content. This methodology enables the creation of the \instruct{} instruction dataset, retaining raw document knowledge within dialogue-style interactions. Utilizing this dataset, we fine-tune \gmodel{}, a model designed to transform raw documents into structured multi-turn dialogues, thereby injecting comprehensive domain knowledge into the SFT model for enhanced instruction tuning. This work signifies a stride towards refining the adaptability and effectiveness of LLMs in processing and generating more accurate, contextually nuanced responses across various fields.
\end{abstract}

\section{Introduction}

The evolution of large language models (LLMs), including advancements seen in the generative pre-trained Transformer (GPT) series by OpenAI, marks a significant milestone in the field of natural language processing (NLP). A pivotal strategy in their development has been instruction-tuning, which leverages human-curated prompts, feedback, and benchmark datasets to tailor LLMs' adaptability to specific domains, such as complex reasoning \cite{cot,xcot,DBLP:journals/corr/abs-2402-11100} and coding \cite{code_llama,starcoder}. Instruction tuning \cite{self_instruct} emerges as an innovative approach, creating new tasks with bespoke instructions, thus enhancing model performance and cost-effectiveness. The diversity and scope of instruction data are critical for the model's ability to generalize and excel in previously unseen tasks, underpinning the continuous advancement and specialization of LLMs in various fields.

\begin{figure}[t]
\begin{center}
	\includegraphics[width=1.0\columnwidth]{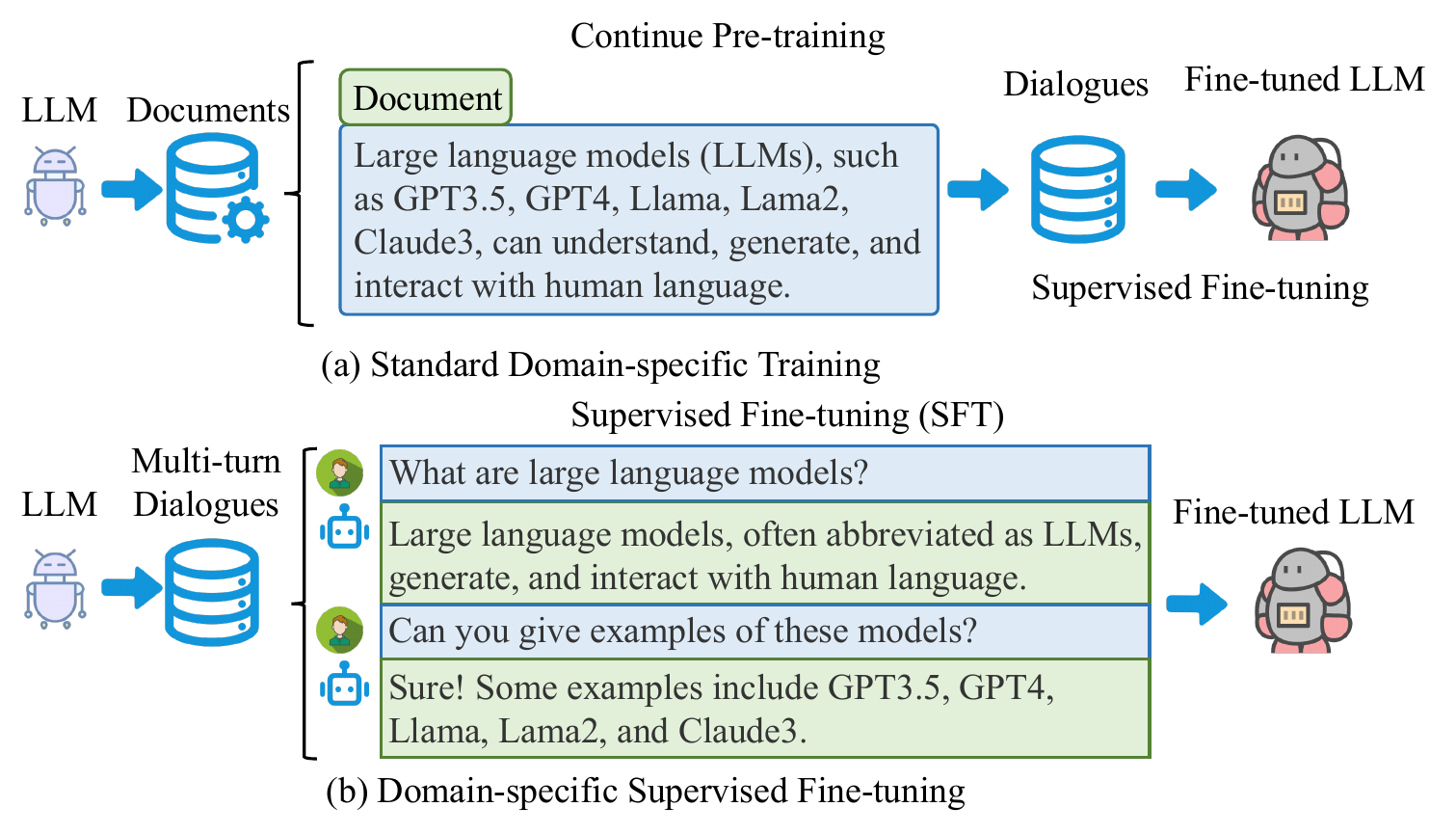}
	\caption{Comparison of standard domain-specific training with our proposed \ourmethod{}. }
	\label{fig:intro}
	\vspace{-10pt}
\end{center}
\end{figure}

Instruction tuning is a groundbreaking technique designed to fine-tune these powerful LLMs for better task-specific performance without intensive retraining on massive datasets, which delicately recalibrates the response of LLMs by giving them instructions or prompts that are carefully crafted to elicit more accurate, contextually appropriate, or nuanced outputs. Many recent works \cite{self_align,wizardmath} try to synthesize instructions, input, and output samples from an LLM and then filter invalid or similar ones. However, these methods focus on generating diverse single-turn dialogues based on the query or response. In \autoref{fig:intro}, continuing pre-training with raw documents and instruction tuning with human-annotated SFT data can inject domain-specific knowledge, but the process requires two-stage training and large-scale data. \textit{Therefore, How to produce reasonable multi-turn dialogues for instruction turning to inject the knowledge of raw documents into LLMs only with instruction tuning.}

In this paper, we propose a framework to construct the \textbf{S}upervised fine-tuning dataset from the \textbf{R}aw documents (\textbf{\ourmethod{}}), which leverages the \textbf{C}hain of \textbf{D}ialogue logic (\cod{}) to guide the LLM to create the reasonable knowledge-intensive multi-turn dialogues for instruction tuning. Specifically, we collect the existing documents from open-source dataset \cite{squadv2,refgpt} and crawl the domain-specific documents from websites (in the field of the artifacts) to create a knowledge-intensive benchmark \benchmark{}, comprised of three tasks, including Wikipedia (English), Science (Chinese), and Artifacts (Chinese). Then, we introduce the chain of dialogue logic to first decide the type (e.g. Opinion Exchange Q\&A, Informational Q\&A, or Task-oriented Q\&A) of the current turn in multi-turn dialogue. subsequently, we prompt the LLM to generate key phrases to search relevant spans for response generation. In this way, we successfully create the instruction dataset \instruct{}, which can keep the knowledge of the raw documents as much as possible in the style of the dialogue. Based on the raw documents, \instruct{}, we can fine-tune \gmodel{} based on open-source LLMs, which aims at converting raw documents into multi-turn dialogues. Finally, we can use \gmodel{} to generate the multi-turn SFT data to inject the knowledge into the SFT model.  

Extensive experiments of \ourmethod{} are evaluated on our created benchmark \benchmark{}. The results demonstrate that the synthetic instruction corpora can effectively inject the knowledge of raw documents into the SFT model, notably getting excellent performance under multiple evaluation metrics. The fine-tuned generator \gmodel{} further verifies the effectiveness of \cod{}, leading to a reasonable and natural multi-turn dialogue. The contributions in this work are summarized as follows:
\begin{itemize}
    \item We propose the chain of dialogue logic (\cod{}), which ingeniously guides LLMs to produce reasonable and natural knowledge-intensive multi-turn dialogues for instruction tuning. It allows the LLMs to generate dialogues that are coherent and contextually relevant but also embed rich, domain-specific knowledge into these conversations. 
    \item We create a knowledge-intensive benchmark (\benchmark{}) to facilitate the training and evaluation of the proposed methods, this work contributes a comprehensive knowledge-intensive benchmark, \benchmark{}, covering a diverse range of topics—including Wikipedia (English), Science (Chinese), and Artifacts (Chinese)—\benchmark{} serves as a vital resource for assessing the effectiveness of \cod{} and the overall framework in handling complex, knowledge-driven dialogue tasks.
    \item The work outlines the creation of \instruct{}, a synthetic instruction dataset that retains an extensive amount of knowledge from the raw documents in a dialogue format, which is used to fine-tune an open-source LLM, referred to as \gmodel{}, which is specifically designed to transform raw documents into cohesive multi-turn dialogues. The experimental results from evaluating \ourmethod{} on the \benchmark{} benchmark demonstrate that this synthetic instruction approach is highly effective in enhancing the SFT model, enabling it to excel across various performance metrics. The fine-tuned generator \gmodel{} is further proof of the efficiency of the \cod{} methodology, producing dialogues that are both logical and rich in domain-specific knowledge.
\end{itemize}

\begin{figure*}[t]
    \centering
    \includegraphics[width=0.75\textwidth]{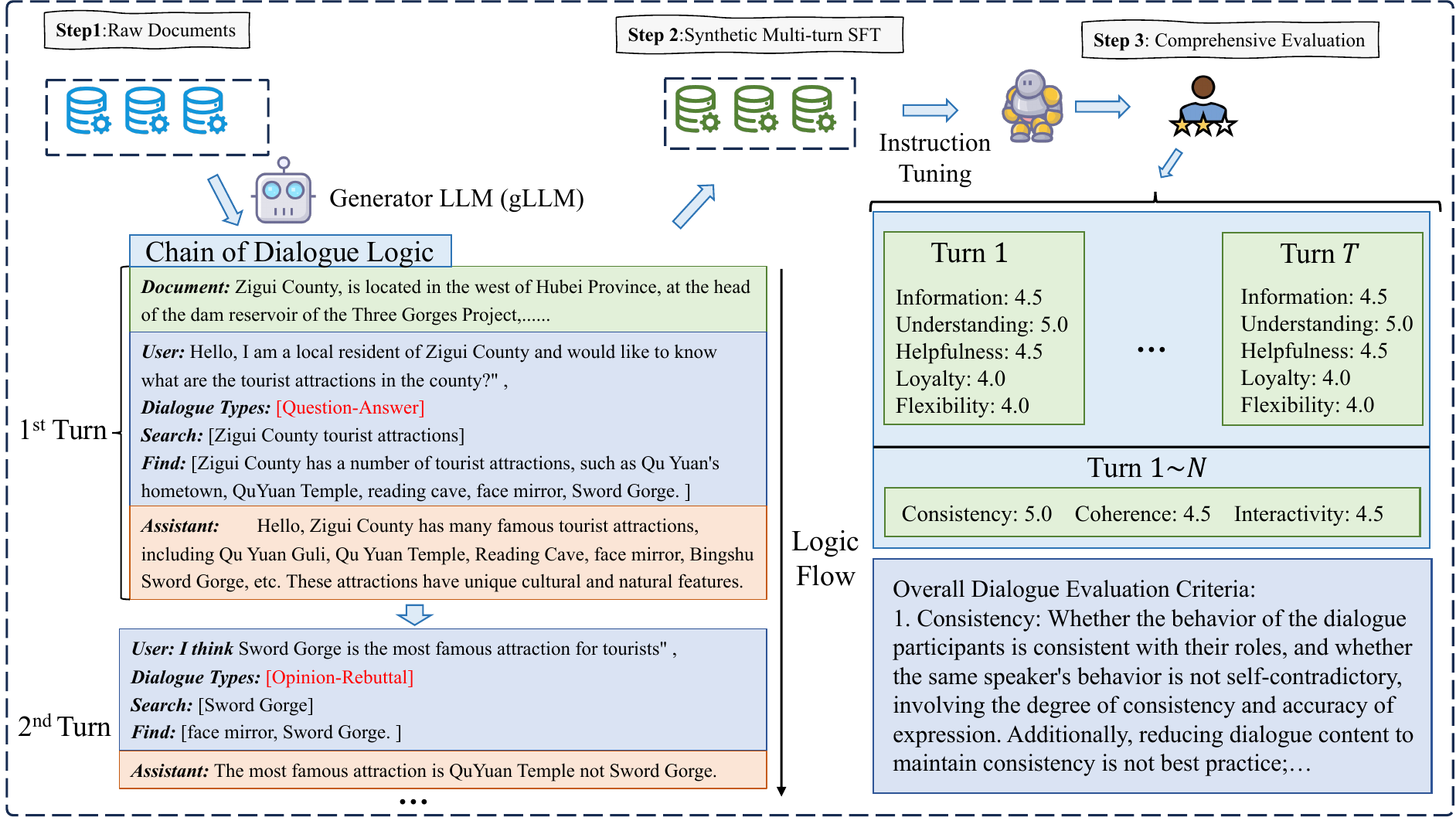}
    \caption{Framework of \ourmethod{}.}
    \label{fig:framework}
    \vspace{-10pt}
\end{figure*}
\section{Problem Definition}
\paragraph{Pre-training} During pre-training, the LLM learns to predict the next word in a document based on the prefix context, enabling the LLM to capture linguistic patterns, syntax, semantics, and even some world knowledge embedded in the text. Given the raw document $x \in D_{d}$, the pre-training can be described as:
\begin{MiddleEquation}
\begin{align}
    P(x) = \prod_{k=1}^{K}P(x_{k}|x_{:k-1})
    \label{pretraining}
\end{align}
\end{MiddleEquation}where $x$ is the document with $K$ tokens.
\paragraph{Instruction Tuning} Supervised fine-tuning (SFT) is the process of taking a pre-trained model, like GPT, and adapting it to a specific task by training on a labeled dataset, which is typically much shorter than the initial pre-training phase since the model has already learned a wealth of language features. Given the query and response $(q,a) \in D_{s}$, the single-turn instruction tuning can be described as:
\begin{MiddleEquation}
\begin{align}
    P(a) = \prod_{i=1}^{n}P(a_{i}|a_{:i-1},q;\mathcal{M})
    \label{instruct_tuning}
\end{align}
\end{MiddleEquation}where $q$ is the query, $a$ is the response, $a_{:i-1}$ is the previous context from the $1$-th to the $i$-$1$-th token.

\paragraph{Instruction Tuning with Raw Text}
If we only have the dataset $D_{d}$ with a limited number of raw documents, which are unable to support the domain-specific pre-training and following SFT, we can create muli-turn dialogues from $D_{d}$ to inject domain-specific knowledge into LLMs by instruction tuning. We create the multi-turn dialogue $\{q^{(t)},a^{(t)}\}_{t=1}^{T} \in D_{s}$ from $D_{d}$ with raw documents to fine-tune the LLM $\mathcal{M}$ as:
\begin{MiddleEquation}
\begin{align}
    P(\{a^{(t)}\}_{t=1}^{T}) = \prod_{t=1}^{T}P(a^{(t)}|a^{(:t-1)},q^{(:t)};\mathcal{M})
    \label{instruct_tuning_with_raw}
\end{align}
\end{MiddleEquation}where $a^{(t)}$ and $q^{(t)}$ are $t$-th turn of the dialogures. $a^{(:t-1)}$ and $q^{(t)}$ are previous $t-1$ queries and $t$ answers.

\section{Benchmark}
\begin{table}
\centering
\resizebox{1.0\columnwidth}{!}
{\begin{tabular}{@{}l|ccc@{}}
\toprule 
            & \textbf{En} & \textbf{Zh} & \textbf{Zh} \\
\midrule
Type & Wikipedia       & Science   & Artifacts      \\
Source   & Squad V2      & RefGPT   & Website Crawler     \\
\#Doc Size & 10989     & 10000   & 6817    \\
\#Instruction Size & 10428     & 9540   & 6152     \\ 
\#Test Size & 1041     & 955   & 615     \\ \midrule
Max Doc Characters  & 4001 & 1526 & 1307 \\
Avg Doc Characters & 1265 & 272 & 584 \\
Max Instruction Characters  & 14009 & 4866 & 5445 \\
Avg Instruction Characters & 7878 & 3656 & 3922 \\ \midrule
\end{tabular}}
\caption{Data statistics. ``\#'' denotes the number of data examples. The maximum and average characters refer to the maximum and average characters of a single document or instruction dataset example.}
\label{tab:data_statistics}
\vspace{-10pt}
\end{table}

\subsection{Data Collection \& Statisitcs}
In \autoref{tab:data_statistics}, our created benchmark comprises three distinct sections categorized by topics. we create $10,428$ English instruction instances and a test size of $1,041$ items from the documents of the Squad V2.0 dataset. Then, the science dataset contains $9,540$ instruction instances and 955 test items from $10,000$ documents. The dataset in Artifacts utilizes data extracted from a website crawler, comprising $6,152$ instructions, and $615$ test items from $6,872$ documents. This dataset reflects a diverse collection strategy, spanning different languages (English and Chinese), content types (General and Artifacts), and sources (Squad V2, RefGPT, and a website crawler), aiming to provide comprehensive resources for domain-specific research.


\subsection{Evaluation Metric}
\label{ssec:eval_metric}
To evaluate the generated multi-turn dialogues, we design the following metrics using the LLM (GPT-4) with the score range (\{1: `very bad', 2: `bad', 3: `neutral', 4: `good', 4: `very good' \}).

\paragraph{Informativeness (Info):} \textbf{Info} is used to accurately whether the query articulates their problem or viewpoint, including the amount of key information, word count, and precision of expression.

\paragraph{Understanding (US):} \textbf{US} is used to evaluate the relevance between queries and the corresponding responses. For each turn of the dialogue, we calculate the score of US and average them to get a final score.

\paragraph{Usefulness (UF):} \textbf{UF} is introduced to decide whether the responses help the user solve their problem or confirm or correct their viewpoints (this may be based on facts or external documents). 

\paragraph{Fidelity (FD):}  \textbf{FD} judges where the response involves factual document information and factual knowledge used in the document.

\paragraph{Flexibility (FL):} For the dialogue without the specific knowledge, \textbf{FL} is used to judge whether the response can correctly handle the query using the common knowledge in LLMs.

\paragraph{Consistency (CS):} Whether the behavior of conversation participants aligns with their roles and the behavior of the same speaker is consistent, involving the consistency and accuracy of expression. 

\paragraph{Cohesion (CO):}: Whether transitions in the conversation, topic deepening, and switches between topics are natural. 

\paragraph{Interactivity (IA):} Whether the user and assistant express and share emotions during the exchange, including the accuracy of emotional expression and accurate understanding of emotions.

\paragraph{Coverage Rate (CR):} For multi-round dialogs constructed on the basis of documents, \textbf{CR} is used to evaluate the proportion of content covered by the dialogs over the content of the documents.

\begin{table*}
\centering
\resizebox{1.0\textwidth}{!}
{\begin{tabular}{@{}l|c|ccccc|cccc|c}
\toprule 
\textbf{Models}          & \textbf{LLM}  & \textbf{Info (Q)} & \textbf{UD (R)} & \textbf{UF (R)} &\textbf{FD (R)} & \textbf{FL (R)} & \textbf{CS} & \textbf{CO} & \textbf{IA} & \textbf{CR} & Avg.\\
\midrule
Direct & GPT-3.5 & 3.15    & 3.60   & 3.60 & 3.60   & 3.15   & 3.57  & 3.57   & 3.12   & 59.18 & 2.96 \\
CoT \cite{cot} & GPT-3.5 & 2.93    & 3.55   & 3.55 & 4.06   & 3.70   & 3.58  & 3.64   & 3.88   & 46.04 & 3.46 \\
\cod{} (our method) & GPT-3.5 & \textbf{3.94}    & \textbf{4.32}   & \textbf{4.32} & \textbf{4.32}   & \textbf{4.13}   & \textbf{4.34}  & \textbf{4.34}   & \textbf{4.28}   & \textbf{74.77} & \textbf{4.19}  \\
\cod{} (our method) & \gmodel{} (Llama-3-8B) & 3.86    & 4.30   & 4.30 & 4.21   & 4.01   & 4.29  & 4.18   & 3.97   & 68.41 & 4.06 \\
\cod{} (our method) & \gmodel{} (Qwen-2-7B) & 3.89    & 4.26   & 4.26 & 4.27   & 4.08    & 4.30  & 4.24   & 4.17   & 67.54 & 4.09 \\
\midrule
Direct & Qwen         & 3.57    & 4.23   & 4.23 & 4.23   & 4.20   & 4.12  & 4.30   & 4.02   & 70.62 & 4.04 \\
CoT \cite{cot} & Qwen & 3.50    & 4.18   & 4.18 & 4.18   & 4.16   & 4.12  & 4.26   & 3.96   & 73.67 & 4.02 \\
\cod{} (our method) & Qwen & \textbf{3.95}    & \textbf{4.45}   & \textbf{4.45} & \textbf{4.43}   & \textbf{4.24}   & \textbf{4.47}  & \textbf{4.48}   & \textbf{4.42}   & \textbf{83.03} & \textbf{4.33}  \\ \midrule
Direct & Deepseek & 3.53    & 4.20   & 4.20 & 4.20   & 4.17   & 4.13  & 4.26   & 4.03   & 60.79 & 3.97 \\
CoT \cite{cot} & Deepseek & 3.64    & 4.28   & 4.28 & 4.29   & 4.18   & 4.25  & 4.34   & 4.11   & 70.02 & 4.09 \\
\cod{} (our method) & Deepseek & \textbf{3.95}    & \textbf{4.46}   & \textbf{4.46} & \textbf{4.44}   & \textbf{4.30}   & \textbf{4.49}  & \textbf{4.50}   & \textbf{4.39}   & \textbf{75.32} & \textbf{4.30}  \\

\bottomrule

\end{tabular}}
\caption{Evaluation results of generated multi-turn SFT data in Artifacts.}
\label{tab:k_bench_artifacts}
\vspace{-10pt}
\end{table*}

\section{Framework of \ourmethod{}}
\label{sec:method}
Figure \ref{fig:framework} describes the overall framework of our method.
\subsection{Logic Definition}
To create the multi-turn dialogue, we consider the following characteristic of the multi-turn dialogue:
(1) \textbf{Contextual Relevance}: In multi-turn dialogues, each round of conversation is related to the context before and after it. Each answer is based on the previous round's question, and the next round's question may be based on the last answer. This coherence is a critical characteristic of multi-turn dialogues.
(2) \textbf{Continuity}: In multi-turn dialogues, the exchange between the user and the assistant is continuous, not isolated single questions and answers, but a series of questions and answers.
(3) \textbf{Dialogue Depth}: In multi-turn dialogues, users may delve deeper into a topic with their inquiries or follow up on the responses received with further questions. This requires the assistant to have the capability to manage complex dialogues and understand the depth of the conversation.
(4) \textbf{Topic Transition}: In multi-turn dialogues, users may switch topics during the conversation. This demands the assistant to be flexible in response and able to answer on new topics.
(5) \textbf{Naturalness of Dialogue}: In multi-turn dialogues, the assistant's responses need to be as natural and fluid as possible, making the user feel as though they are communicating with a person, not a machine.

Inspired by these characteristics of the multi-turn dialogue, we pre-define the six logic types of each turn: 
(1) \textcolor{blue!60}{Question-Answer}: This is the most common logical sequence in dialogue, where Character A asks a question and Character B provides an answer. For example, `How old are you?', `I'm 23 years old.'.
(2) \textcolor{blue!60}{Question-Question}: This is a logical process where an intent to ask is completed. Character A asks a relatively vague question, and if Character B needs to clarify the intent of Character A's question, Character B can continue by asking another question. For example, `How old are you?', `Are you asking about my age?'.
(3) \textcolor{blue!60}{Statement-Inquiry}: Character A makes a statement, and Character B asks for or requires more information. For example, "A: I went to the museum today." "Oh, what interesting exhibitions did you see?"
(4) \textcolor{blue!60}{Statement-Explanation}: Character A states a fact, and then Character B explains or elaborates on the information related to that fact. For example, `Yao Ming won the CBA championship in 2002.', `B: He was also named CBA Rebound King three times and Block King, and twice the CBA Dunk King.'.
(5) \textcolor{blue!60}{Opinion-Rebuttal}: Character A presents an opinion, and Character B counters with facts or presents a different viewpoint. For example, `A: I think this movie is really good.', `No, this movie scored a 9.5 rating, it is a good movie.'.
(6) \textcolor{blue!60}{Opinion-Agreement}: Character A expresses an opinion, and Character B either agrees or disagrees based on facts. For example, `A: That singer performed terribly.', `Indeed, the judges gave very low scores.'.

Each link in the dialogue logical chain contains the current turn's dialogue type, progress, the logical process of the dialogue participants, and the purpose of the dialogue, achieving the effect of guiding the generation of multi-turn dialogue.

\begin{table*}
\centering
\resizebox{0.95\textwidth}{!}
{\begin{tabular}{@{}l|c|ccccc|cccc|c}
\toprule 
\textbf{Models} & \textbf{LLM}& \textbf{Info (Q)} & \textbf{UD (R)} & \textbf{UF (R)} &\textbf{FD (R)} & \textbf{FL (R)} & \textbf{CS} & \textbf{CO} & \textbf{IA} & \textbf{CR} & Avg.\\
\midrule
Direct    & GPT-3.5& 3.24    & 3.66   & 3.66 & 3.67   & 3.24   & 3.65  & 3.65   & 2.98   & 21.18 & 3.20  \\ 
CoT \cite{cot} & GPT-3.5 & 2.95    & 3.67   & 3.67 & 4.03   & 3.72   & 3.67  & 3.70   & 3.46   & 19.50 & 3.42 \\

\cod{} (our method)     & GPT-3.5 & \textbf{3.96}    & \textbf{4.39}   & \textbf{4.39} & \textbf{4.39}   & \textbf{4.20}   & \textbf{4.42}  & \textbf{4.42}   & \textbf{4.29}   & 40.55 &  \textbf{4.05}  \\
\cod{} (our method) & \gmodel{} (Llama-3-8B) & 3.92    & 4.36   & 4.36 & 4.27   & 4.16  & 4.37  & 4.36   & 4.25   & 35.42 & 3.98 \\
\cod{} (our method) & \gmodel{} (Qwen-2-7B) & 3.93    & 4.34   & 4.34 & 4.29   & 4.19    & 4.40  & 4.38   & 4.25   & 36.74 & 3.99 \\
\midrule
Direct & Qwen & 3.37    & 4.16   & 4.16 & 4.17   & 4.13   & 4.07  & 4.16   & 3.91   & 33.07 & 3.75 \\
CoT \cite{cot} & Qwen & 3.42    & 4.16   & 4.16 & 4.17   & 4.13   & 4.12  & 4.16   & 3.95   & 35.07 & 3.78 \\
\cod{} (our method) & Qwen& \textbf{3.80}    & \textbf{4.36}   & \textbf{4.36} & \textbf{4.33}   & \textbf{4.20}   & \textbf{4.36}  & \textbf{4.40}   & \textbf{4.30}   & \textbf{51.69} & \textbf{4.07}  \\ \midrule
Direct & Deepseek & 3.53    & 4.11   & 4.11 & 4.16   & 4.11   & 4.10  & 4.18   & 4.02   & 21.71 & 3.71 \\
CoT \cite{cot} & Deepseek & 3.51    & 4.19   & 4.19 & 4.20   & 4.13   & 4.16  & 4.22   & 4.01   & 34.24 & 3.81 \\
\cod{} (our method) & Deepseek & \textbf{3.81}    & \textbf{4.38}   & \textbf{4.38} & \textbf{4.34}   & \textbf{4.20}   & \textbf{4.40}  & \textbf{4.41}   & \textbf{4.34}   & 36.29 & \textbf{4.00}  \\ \midrule
RefGPT \cite{refgpt} & - & 3.66    & 4.33   & 4.33 & 4.28   & \textbf{4.20}   & 4.23  & 4.31   & 4.04 & 45.57 & 3.96 \\
\bottomrule
\end{tabular}}
\caption{Evaluation results of generated multi-turn SFT data in Science. The original dialogues from RefGPT~\cite{refgpt} are evaluated using the same merics and scoring methods for comparison.}
\label{tab:k_bench_science}
\vspace{-10pt}
\end{table*}

\subsection{Chain of Dialogue Logic}
The chain of dialogue logic (DTC) encourages large language models (LLMs) to mimic the human thought process between rounds of dialogue, which includes identifying the type of dialogue, searching for relevant information, and discovering pertinent details, aiming to guide the generation of dialogue between turns. The design of the DTC format in this paper is inspired by chain of thought (CoT) and React. CoT is an improved prompting strategy intended to enhance the performance of LLMs in complex reasoning tasks such as arithmetic reasoning, common sense reasoning, and symbolic reasoning. React represents a new prompting paradigm based on repeated thought-action observation cycles until the current knowledge suffices to derive an answer, employed to facilitate reasoning and action within language models to tackle general tasks. DTC benefits from the ``Thought'' paradigm of Dialogue-Search-Find, generating responses in dialogue that are more accurate and richer compared to CoT and React, as the Dialogue Types component leads the type of dialogue response, determining the direction for generating answers. Meanwhile, the Search-Find component can introduce factual knowledge into the dialogue, enhancing answer accuracy. Furthermore, unlike the simple prompt construction approach of input-output pairs, DTC incorporates intermediate reasoning steps, guiding the model towards the direction of the final output.

\subsection{\gmodel{}}
Given the synthetic multi-turn dialogue $\{q^{(t)},a^{(t)}\}_{t=1}^{T} \in D_{s}$ generated by the teacher LLM (e.g. GPT-4), we can fine-tune the open-source LLMs to obtain the generator \gmodel{} based on Qwen-2-7B and Llama-3-8B. To reduce the cost, we can use the \gmodel{} with fewer parameters to inference more samples by adjusting the sampling temperature and sampling. In our work, we create the sampled dataset $\{q^{(t)}_{g},a^{(t)}_{g}\}_{t=1}^{T} \in D_{s}$ from \gmodel{} to augment the original dataset.

\section{Experiments}

\begin{table*}
\centering
\resizebox{0.95\textwidth}{!}
{\begin{tabular}{@{}l|c|ccccc|cccc|c}
\toprule 
\textbf{Models}        & \textbf{LLM} & \textbf{Info (Q)} & \textbf{UD (R)} & \textbf{UF (R)} &\textbf{FD (R)} & \textbf{FL (R)} & \textbf{CS} & \textbf{CO} & \textbf{IA} & \textbf{CR} & Avg.\\
\midrule
Direct              & GPT-3.5 & 3.26    & 3.64   & 3.64 & 3.67   & 3.31   & 3.59  & 3.59   & 3.16   & 22.66 & 3.22 \\ 
CoT \cite{cot}      & GPT-3.5 & 3.24    & 3.95   & 3.95 & 4.10   & 3.96   & 3.96  & 4.01   & 3.73   & 28.73 & 3.59 \\
\cod{} (our method) & GPT-3.5 & \textbf{3.87}    & \textbf{4.30}   & \textbf{4.30} & \textbf{4.34}   & \textbf{4.11}   & \textbf{4.35}  & \textbf{4.35}   & \textbf{4.30}   & \textbf{73.03} & \textbf{4.17} \\ 
\cod{} (our method) & \gmodel{} (Llama-3-8B) & 3.86    & 4.28   & 4.28 & 4.25   & 4.07   & 4.32  & 4.32   & 4.01   & 66.28 & 4.07 \\
\cod{} (our method) & \gmodel{} (Qwen-2-7B) & 3.87    & 4.26   & 4.26 & 4.29   & 4.10    & 4.34  & 4.34   & 4.10   & 69.44 & 4.11 \\
\midrule
Direct & Qwen & 3.49    & 4.15   & 4.15 & 4.16   & 4.12   & 4.08  & 4.21   & 4.01   & 43.92 & 3.83 \\
CoT \cite{cot} & Qwen & 3.58    & 4.25   & 4.25 & \textbf{4.26}   & 4.10   & 4.24  & 4.29   & 4.06   & 60.29 & 4.00 \\
\cod{} (our method) & Qwen& \textbf{3.75}    & \textbf{4.36}   & \textbf{4.36} & 4.20   & \textbf{4.16}   & \textbf{4.36}  & \textbf{4.38}   & \textbf{4.31}   & \textbf{68.01} & \textbf{4.14}  \\
\midrule
Direct & Deepseek & 3.49    & 4.11   & 4.10 & 4.12   & 4.10   & 4.11  & 4.18   & 4.05   & 33.79 & 3.77 \\
CoT \cite{cot} & Deepseek & 3.64    & 4.32   & 4.32 & 4.29   & 4.10   & \textbf{4.31}  & 4.33   & 4.09   & 62.25 & 4.05 \\
\cod{} (our method) & Deepseek & \textbf{3.69}    & \textbf{4.33}   & \textbf{4.33} & \textbf{4.27}   & \textbf{4.17}   & \textbf{4.31}  & \textbf{4.34}   & \textbf{4.29}   & \textbf{65.13} & \textbf{4.10}  \\
\bottomrule
\end{tabular}}
\caption{Evaluation results of generated multi-turn SFT data in Wikipedia. }
\label{tab:k_bench_wikipedia}
\vspace{-15pt}
\end{table*}

We conduct three types of experiments to evaluate the ability and effectiveness of our proposed \ourmethod{} framewor to inject document knowledge into multi-turn dialogues:
\begin{itemize}
\item \textbf{\benchmark{} Evaluation}: we assess the quality of \benchmark{} by conducting a comprehensive assessment involving both automated and human evaluation methods, subsequently
\item \textbf{\gmodel{} Evaluation}: we examine how effectively \gmodel{} can generate multi-turn supervised fine-tuning (SFT) data with document knowledge injected
\item \textbf{Fine-tuned LLM Evaluation}: we aim to ensure that the answers produced by LLMs trained on the \gmodel{}-generated multi-turn dialogue data also exhibit a high degree of coherence, contextual relevance, and factual accuracy.
\end{itemize}

\subsection{Experimental Setup}

\paragraph{Base LLMs}
For constructing \instruct{} and \benchmark{}, we employ GPT-3.5 Turbo~\cite{chatgpt}, Deepseek V2 Chat~\cite{deepseek_v2}, and Qwen-2-72B-Instruct~\cite{Qwen}.
Based on the pre-trained LLMs with different model sizes, we utilize Llama-2-14B, Llama-3-8B\footnote{\url{https://github.com/meta-llama/llama}}, and Qwen-2-1.5B/7B\footnote{\url{https://github.com/QwenLM/Qwen2}} as the foundation models to construct \gmodel{}.
For model-based evaluation, we employ GPT-4 Turbo (GPT-4)~\cite{gpt4} for high standard.

\paragraph{Evaluation Benchmark}
We adopt both \instruct{} and \benchmark{} created by GPT-3.5 for training and evaluation, though more variants have been created for experimental purposes.
Data statistics can be found in Table \ref{tab:data_statistics}.

\paragraph{Implemetation Details}
Models from all experiments are trained for 2 epochs with a cosine scheduler, starting at a learning rate of 2e-5 (3\% warmup steps). We use AdamW~\citep{adamw} as the optimizer and a batch size of 512 (max length $4096$).
We adopt the evaluation metric introduced in Section \ref{ssec:eval_metric}.

\subsection{Main Results}
\paragraph{\benchmark{} Evaluation}
To verify the quality of of \benchmark{} and further assess the effectiveness of the proposed \cod{} compared to CoT and direct-instruction prompting approaches, we employ GPT-4 Turbo as the judge to evaluate the \benchmark{}. 
GPT-4 Turbo was tasked with scoring each dialogue turn based on the evaluation metrics and scoring standard described in \ref{ssec:eval_metric}.
We adopt GPT-3.5 Turbo (GPT 3.5), Deepseek V2 Chat (Deepseek), and Qwen-2-72B-Instruct (Qwen) as different variants of raw text (documents) to multi-turn SFT data generators.
Results of experiments on the three different types of SFT data generated: artifacts, science, and wikipedia are shown in Table \ref{tab:k_bench_artifacts}, \ref{tab:k_bench_science}, and \ref{tab:k_bench_wikipedia} repectively.
By examing the results, we find that the LLMs with our proposed \cod{} design outperforms other models on all metrics, except for the coverage rate (CR) metric evaluated on the original dialogues created by RefGPT~\cite{refgpt}, using the same evaluation standards in Table \ref{tab:k_bench_science}.
\citet{refgpt} designed specific knowledge-injecting methods to ensure the dialogue data they construct include most of the facts from the documents, while our method not only ensures high factual accuracy, but also exhibits a high degree of coherence and contextual relevance.
Moreover, the SFT data generated by Qwen with the CoD design still outperforms RefGPT on the CR metric, which demonstrates the potential of our method reaching high limits on different backbome LLMs.
Since the \instruct{} dataset proposed for training \gmodel{} is created using the same methods as creating \benchmark{}, the evaluation results also indicate that the \instruct dataset exhibits high quality across different metrics.

\paragraph{\gmodel{} Evaluation}
In Table \ref{tab:k_bench_artifacts}, \ref{tab:k_bench_science}, and \ref{tab:k_bench_wikipedia}, we also present results of \gmodel{} initialized on both Llama-3-8B and Qwen-2-7B.
In detail, we evaluate on the SFT data generated by these two \gmodel{}s using the same metrics and standards introduced in ``\benchmark{} Evaluation''.
Evaluation results of the SFT data generated by GPT-3.5 only slightly is only slightly higher than the data generated by \gmodel{}, which indicates that the \gmodel{}s intialized on both LLMs obtained a high degree of data generation ability from GPT-3.5 with CoD.
By leveraging the chain of dialogue logic and knowledge-intensive documents, the SFT data produced by \gmodel{} surpasses traditional retrieval-based and direct response methods, establishing new high ground for knowledge-intensive instruction tuning in large language models.

\paragraph{Fine-tuned LLM Evaluation}
To evaluate the effectiveness of LLMs fine-tuned on the SFT data produced by \gmodel{}, we conduct experiments comparing different generators, fine-tuned LLMs, and model settings on 3 dataset sources seperately: artifacts, RefGPT and SquadV2.
We exam answers produced by Qwen-2-7B and Llama3-8B fine-tuned on multi-turn dialogue data generated by GPT-3.5 and \gmodel{}, respectively.
To further investigate the effectiveness of \cod{}, we compare LLMs fine-tuned on the training data generated by generators with and without \cod{} (refers as Direct model in \ref{tab:sft_model}) guidance.
Experiment results are presented in Table \ref{tab:sft_model}.
We find that all \cod{} models outperform direct models without \cod{}, demonstrating the effectiveness our chain-of-logic design.
The LLMs trained on data produced by \gmodel{} generators only slighly underperform those trained on data produced by GPT-3.5. 
This also matches the results from the \gmodel{} evaluation experiment, where \gmodel{} acquired most of the dialogue generation abilities from its upper-boundary GPT-3.5.

\subsection{Analysis}

\begin{table*}
\centering
\resizebox{0.95\textwidth}{!}
{\begin{tabular}{@{}l|ccc|ccccc|cccc|c}
\toprule 
\textbf{Models}       & \textbf{Dataset}  & \textbf{Fine-tuned LLM} & \textbf{Generator} & \textbf{Info (Q)} & \textbf{UD (R)} & \textbf{UF (R)} &\textbf{FD (R)} & \textbf{FL (R)} & \textbf{CS} & \textbf{CO} & \textbf{IA} & \textbf{CR} & Avg.\\
\midrule
Direct  & Artifacts & Qwen-2-7B & GPT-3.5   & 3.90    & 4.30   & 4.28 & 3.89   & 4.06   & 4.45  & 4.45   & 4.44   & 67.61 & 4.12\\ 
Direct   & Artifacts & Llama-3-8B & GPT-3.5 & 3.38    & 4.09   & 4.05 & 3.67   & 3.97   & 4.19  & 4.16   & 3.80   & 59.47 & 3.80 \\
\cod{}  & Artifacts & Qwen-2-7B & GPT-3.5   & \textbf{3.93}    & 4.31   & \textbf{4.30} & 3.98   & \textbf{4.11 }  & \textbf{4.45}  & 4.45   & 4.45   & \textbf{80.02} & \textbf{4.22}\\
\cod{}   & Artifacts & Llama-3-8B & GPT-3.5 & 3.84    & 4.03   & 4.03 & \textbf{4.08}   & 4.07   & 4.22  & 4.23   & 4.21   & 72.49 & 4.03\\
\cod{} & Artifacts & Qwen-2-7B & \gmodel{} & 3.89    & \textbf{4.32}   & \textbf{4.30} & 3.91   & 4.04   & 4.42  & \textbf{4.43}   & \textbf{4.45}   & 75.33 & 4.16\\
\cod{} & Artifacts & Llama-3-8B & \gmodel{} & 3.75    & 3.98   & 3.98 & 3.89   & 4.01   & 4.19  & 4.21   & 4.19   & 64.57 & 3.93\\
\midrule
Direct   & RefGPT & Qwen-2-7B & GPT-3.5  & 3.85    & 4.21   & 4.20 & 3.65   & 3.95   & 4.36  & 4.36   & 4.38   & 31.41 & 3.83\\
Direct   & RefGPT & Llama-3-8B & GPT-3.5 & 3.54    & 3.88   & 3.88 & 3.57   & 3.77  & 3.94  & 3.94   & 3.95   & 28.55 & 3.54\\
\cod{}   & RefGPT & Qwen-2-7B & GPT-3.5  & \textbf{3.87}    & \textbf{4.24}   & \textbf{4.24} & \textbf{3.76}   & \textbf{4.04}   & \textbf{4.39}  & \textbf{4.40}   & 4.40   & \textbf{57.23} & \textbf{4.02}\\
\cod{}   & RefGPT & Llama-3-8B & GPT-3.5 & 3.60    & 3.97   & 3.97 & 3.69   & 3.87   & 3.99  & 4.00   & 3.99  & 52.84 & 3.74\\
\cod{}  & RefGPT & Llama-3-8B & \gmodel{} & 3.50    & 3.91   & 3.91 & 3.66   & 3.85   & 3.93  & 3.92   & 3.98   & 45.11 & 3.65\\
\cod{}   & RefGPT & Qwen-2-7B & \gmodel{} & 3.81    & 4.22   & 4.22 & 3.69   & 3.95   & 4.34  & 4.34   & \textbf{4.41}   & 42.38 & 3.89\\
\midrule
Direct    & SquadV2 & Qwen-2-7B & GPT-3.5 & 3.86    & \textbf{4.10}   & \textbf{4.10} & 3.64   & 3.92   & \textbf{4.23}  & \textbf{4.20}   & 4.04   & 53.14 & 3.86\\
Direct  & SquadV2 & Llama-3-8B & GPT-3.5 & 3.80    & 3.56   & 3.56 & 3.26   & 3.40   & 3.91  & 3.82   & 3.76   & 46.32 & 3.48\\
\cod{}    & SquadV2 & Qwen-2-7B & GPT-3.5 & \textbf{3.87}    & \textbf{4.10}   & \textbf{4.10} & 3.67   & \textbf{3.95}   & \textbf{4.23}  & \textbf{4.20}   & \textbf{4.07}   & \textbf{61.84} & \textbf{3.92}\\
\cod{}  & SquadV2 & Llama-3-8B & GPT-3.5 & 3.76    & 3.79   & 3.79 & 3.32   & 3.49   & 4.00  & 3.88   & 3.84   & 48.98 & 3.59\\
\cod{}   & SquadV2 & Llama-3-8B & \gmodel{} & 3.58    & 3.75   & 3.75 & 3.30   & 3.49   & 3.89  & 3.74   & 3.86   & 58.01 & 3.58\\
\cod{}   & SquadV2 & Qwen-2-7B & \gmodel{} & 3.80    &  4.01  & 4.01 & \textbf{3.74}   & 3.94   & 4.27  & 4.17   & 4.12   & 51.73 & 3.84\\
\midrule
\end{tabular}}
\caption{Evaluation results of different LLMs fine-tuned on the generated data. Direct denotes the generator is trained with a direct instruction to generate dialogues, while \cod{} refers to training with the chain of dialogue logic design. Fine-tuned LLM is the dialogue model trained on the generated data.}
\label{tab:sft_model}
\vspace{-10pt}
\end{table*}

\paragraph{Human Evaluation}
To ensure the robustness of the GPT-4-based evaluation, we sampled 100 examples from \benchmark{} created by GPT-3.5 and conducted human evaluation.
The human evaluators were provided with the evaluation metrics and detailed instructions to score each dialogue turn from 1 to 5.
The scores were aggregated to provide an overall assessment of the dataset quality.
The evaluators were encouraged to provide qualitative feedback on the naturalness and factual accuracy of the dialogues.
Additionally, we performed a correlation analysis between the scores assigned by GPT-4 and human evaluators.
Specifically, we adopt Pearson and Spearman correlation coefficients to measure the association between the automated and human evaluations.
The Pearson and Spearman correlation coefficients between GPT-4 and human evaluations were 0.89 and 0.87, respectively, indicating a strong alignment between automated and human assessments.

\begin{figure}[t]
    \centering
    \includegraphics[width=1.0\columnwidth]{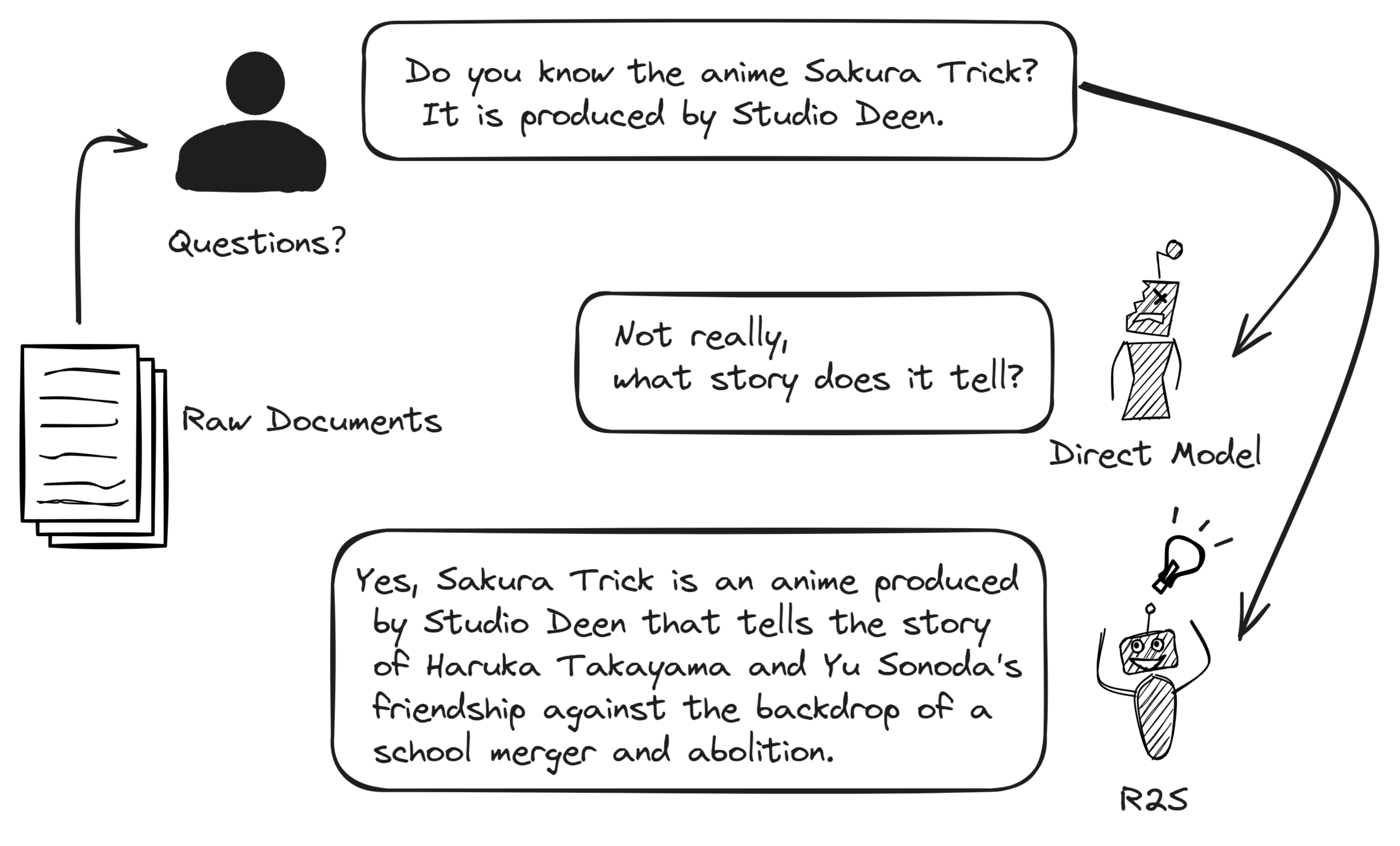}
    \caption{Comparison of dialogues responses generated by direct models and \ourmethod{}}
    \label{fig:dia_compare}
    \vspace{-10pt}
\end{figure}

\paragraph{The effectiveness of \ourmethod{}}
Figure \ref{fig:dia_compare} is a case study comparing dialogue responses generated by direct models and \ourmethod{}.
By integrating the chain of dialogue logic, \gmodel{} effectively maintains factual accuracy and produces coherent, contextually relevant dialogues that mimic human-like interactions.
Such approach not only improves the performance of the generated dialogues but also provides a robust framework for instruction tuning of large language models using raw text documents.
Owing to the \cod{} design: 1) the dialogues generated by \gmodel{} were significantly more informative and detailed, providing users with comprehensive responses, 2) the understanding and coherence of the dialogues were markedly improved, with responses that logically followed the user's queries, and 3)the dialogues exhibited higher loyalty to the reference documents, ensuring factual accuracy and reducing the occurrence of hallucinations.

\section{Related Work}
\paragraph{Large Language Model}
Large language models (LLMs) \cite{llama,llama2,gptq,Qwen,glm,code_llama}, leveraging the Transformer architecture, represent a significant leap in natural language processing (NLP)~\cite{DBLP:conf/acl/LiZLLLSWLCZ22,DBLP:journals/corr/abs-2311-11268,DBLP:journals/corr/abs-2405-12819}. LLMs undergo rigorous training on extensive textual datasets, enabling them to grasp a wide range of linguistic nuances and contexts. LLMs follow a two-stage process involving pre-training on large-scale corpora followed by instruct tuning for specific tasks, significantly improving performance across downstream understanding and generation challenges. Notably, the GPT series, starting from GPT-1 and evolving through GPT-4, \cite{gpt,gpt-neo,gpt3,gpt4} showcases a progressive increase in model complexity and capacity, with GPT-3 comprising a staggering 175 billion parameters. The introduction of instruction tuning further amplifies the capabilities of LLMs, unlocking emergent abilities for intricate reasoning tasks, such as math and code. LLMs with instruction tuning garner attention from researchers and make significant impacts across various industry scenarios.

\paragraph{Instruction Tuning}
LLMs refine their ability to follow and understand user commands more accurately fine-tuned on an instruction dataset \cite{instructgpt,Instruction_Tuning, DBLP:journals/corr/abs-2404-04925}, consisting of various instructions and their corresponding desired outputs.
Early research in constructing conversational datasets largely relies on manually annotated sets (e.g. QuAC \cite{quac} and CoQA \cite{coqa}), but the limited scale and high annotation costs restrict model performance and generalizability. Simulation-based approaches are adopted to generate synthetic dialogues through mimicking user-system interactions, thus reducing dependence on manual annotations. Recent advancements \cite{self_align} emphasize the capabilities of LLMs like GPT-4 in auto-generating extensive, high-quality datasets such as the SODA dataset \cite{soda}, with 1100 million utterances and 3 billion tokens. These works highlight LLMs' data generation prowess and the importance of human intervention in enhancing the accuracy of generated data, for instance, by applying basic, safety, and commonsense filters to GPT-3.5-generated dialogues \cite{llm_human_evaluation,document_grounded_dialogue}. 

\section{Conclusion}
In this paper, we introduce \ourmethod{}, a framework for constructing a supervised fine-tuning (SFT) dataset from raw documents utilizing the chain of dialogue logic (\cod{}) to guide LLMs in creating knowledge-intensive multi-turn dialogues for instruction tuning. By aggregating existing documents from open-source websites/datasets, we establish a comprehensive benchmark, \benchmark{}, featuring topics in Wikipedia (English), Science (Chinese), and Artifacts (Chinese). Utilizing \cod{}, we categorize dialogue turns (e.g., Opinion Exchange Q\&A and Informational Q\&A) and prompt the LLM to identify key phrases for generating relevant responses, thus retaining raw document knowledge within dialogue-style instruction datasets (\instruct{}) and enabling the fine-tuning of open-source LLMs (\gmodel{}) for transforming raw documents into multi-turn dialogues, further enriching the SFT model with domain-specific knowledge. Extensive experiments on \benchmark{} validate the efficacy of \ourmethod{}, showing substantial improvements in performance across various metrics. The SFT generator \gmodel{} confirms the effectiveness of \cod{} by creating reasonable and coherent multi-turn dialogues. 

\clearpage
\section*{Limitations}
Due to limited computing resources, the supervised fine-tuning (SFT) dataset generated by \gmodel{} is evaluated on Qwen-2-7B and Llama-3-8B only, while more backbone large language models should be tested, ensuring the robustness of R2S to inject document knowledge into SFT data. Besides, the large-scale training scenario with more powerful LLMs still requires further exploration.

\section*{Ethical Considerations}
The dataset used for evaluation in this paper is obtained from open data sources and has been manually verified and screened to eliminate any data with ethical risks and sensitive content. This ensures that the content is compliant with existing regulations and laws.


\bibliography{anthology,custom}

\clearpage
\appendix

\end{document}